\documentclass[10pt,twocolumn,letterpaper]{article}

\usepackage{cvpr}              

\usepackage{graphicx}
\usepackage{amsmath}
\usepackage{amssymb}
\usepackage{booktabs}
\usepackage{pifont}
\usepackage{hyperref}

\usepackage[capitalize]{cleveref}
\crefname{section}{Sec.}{Secs.}
\Crefname{section}{Section}{Sections}
\Crefname{table}{Table}{Tables}
\crefname{table}{Tab.}{Tabs.}

\def\cvprPaperID{44} 
\def\confName{CVPR}
\def\confYear{2023}

\begin{document}

\title{Waterfall Transformer for Multi-person Pose Estimation}


\author{Navin Ranjan \hspace{0.75cm} Bruno Artacho \hspace{0.75cm} Andreas Savakis \\
Rochester Institute of Technology\\
Rochester, New York 14623, USA\\
{\tt\small nr4325@rit.edu}\hspace{0.5cm}  {\tt\small bmartacho@mail.rit.edu} \hspace{0.5cm}{\tt\small andreas.savakis@rit.edu}}
\maketitle

\begin{abstract}
We propose the Waterfall Transformer architecture for Pose estimation (WTPose), a single-pass, end-to-end trainable framework designed for multi-person pose estimation. Our framework leverages a transformer-based waterfall module that generates multi-scale feature maps from various backbone stages. The module performs filtering in the cascade architecture to expand the receptive fields and to capture  local and global context, therefore increasing the overall feature representation capability of the network. Our experiments on the COCO dataset demonstrate that the proposed WTPose architecture, with a modified Swin backbone and transformer-based waterfall module, outperforms other transformer architectures for multi-person pose estimation.
\end{abstract}

\section{Introduction}
\label{sec:intro}
Human Pose estimation is a challenging computer vision task with a wide range of practical applications~\cite{zhang2019human},~\cite{tang2018deeply}. 
Deep learning methods based on Convolutional Neural Networks (CNNs) have increased state-of-the-art performance~\cite{wei2016convolutional}, pose~\cite{artacho2021unipose+},\cite{cao2017realtime}. 
Recently, vision transformers~\cite{dosovitskiy2020image}, ~\cite{HRFormer_yuan2021hrformer},~\cite{xu2022vitpose},~\cite{liu2021swin},~\cite{vaswani2017attention},~\cite{DiNAT_hassani2022dilated} have shown excellent performance in computer vision tasks, including pose estimation.

\begin{figure}
  \includegraphics[width=\linewidth]{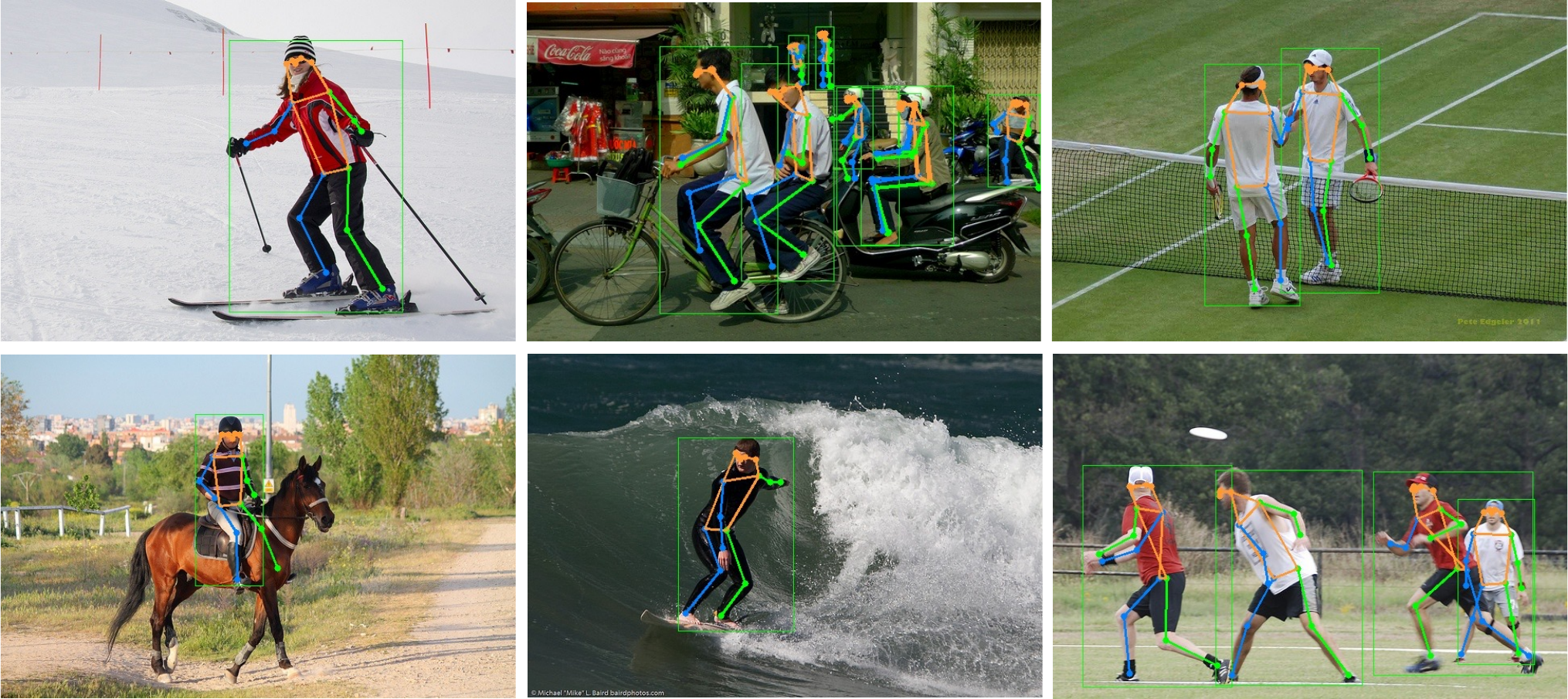}
  \caption{WTPose examples from the COCO dataset.}
  \label{fig:WTPose_examples}
\end{figure}

In this paper, we propose WTPose, a "Waterfall Transformer" architecture in a flexible framework for improved performance over the baseline. Pose estimation examples using WTPose are shown in Figure \ref{fig:WTPose_examples}.
A key feature of our architecture is the integration of our 
multi-scale Waterfall Transformer
Module (WTM) to enhance the performance of vision transformer models, such as the Shifted Window (Swin) transformer \cite{liu2021swin}. We process the feature maps from multiple levels of the backbone through the waterfall branches of WTM. 
The module performs filtering operations based on a dilated attention mechanism to increase the Field-of-View (FOV) and capture both local and global context, leading to  significant performance improvements. The contribution of this paper are the following.
\begin{itemize}
\vspace{-0.1in}
    \item We introduce the novel Waterfall Transformer architecture for pose estimation, a single-pass, end-to-end trainable, multi-scale approach for top-down multi-person 2D pose estimation. 
\vspace{-0.1in}
    \item We propose a waterfall transformer module with multi-scale attention, that employs a dilated attention mechanism enabling a larger receptive field to capture global and local context.
    \vspace{-0.1in}
    \item Our experiments on the COCO dataset demonstrate improved pose estimation performance over comparable transformer methods.
\end{itemize}

\begin{figure*}[t]
\centering
  \includegraphics[width=\linewidth]{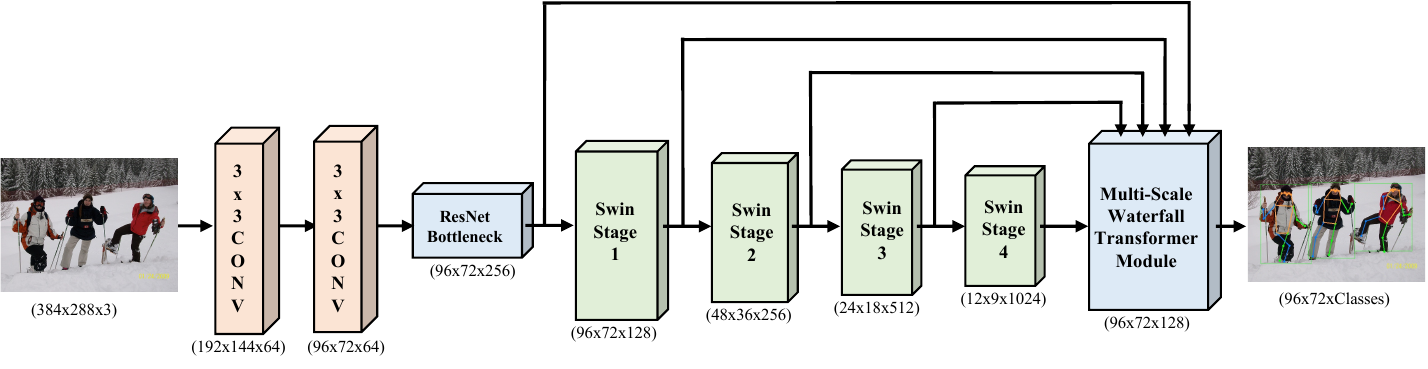}
  \caption{Waterfall transformer framework for multi-person pose estimation. The input color image is fed through the modified Swin Transformer backbone and WTM module to obtain 128 feature channels at reduced resolution by a factor of 4. The decoder module generates K heatmaps, one per joint.}
  \label{fig:WTarchitecture}
\end{figure*}

\section{Related Work}
\label{sec:related_work}

\subsection{CNNs for pose estimation}
With the advancement of deep convolutional neural networks, human pose estimation has achieved superior results~\cite{wei2016convolutional},~\cite{DNN_Pose_6909610},~\cite{artacho2021unipose+},~\cite{cao2017realtime}. The Convolutional Pose Machine (CPM)~\cite{wei2016convolutional} architecture includes multiple stages, producing increasingly refined joint detection. The OpenPose method~\cite{cao2017realtime} included Part Affinity Fields to deal with pose of multiple people in an image. The Stacked Hour-glass network~\cite{newell2016stacked} uses repeated bottom-up and top-down processing with intermediate supervision to process across all scales and capture the best spatial relationship associated with the body for accurate pose estimation. Expanding on the stacked hourglass networks, the multi-context attention approach~\cite{Chu_2017_CVPR} designs Hourglass Residual Units (HRUs) with the goal of generating attentions maps with larger receptive fields at multiple resolutions and with various semantics. Additionally, post-processing with Conditional Random Fields (CRFs) is used to generate locally and globally consistent pose estimates. 


The High-Resolution Network (HRNet) architecture ~\cite{sun2019deep},~\cite{wang2020deep} connects high-to-low sub-networks in parallel,  maintaining high-resolution representations throughout the process, and generating more accurate and spatially precise pose estimates. The Multi-Stage Pose Network~\cite{Li2019RethinkingOM} operates similarly with HRNet~\cite{wang2020deep}, but it employs a cross-stage feature aggregation strategy to propagate information from early stages to the latter ones and is equipped with coarse-to-fine supervision. 

The UniPose+~\cite{artacho2021unipose+}, OmniPose~\cite{OmniPose}, 
and BAPose~\cite{Artacho_WACV2023} methods propose multiple variants of the Waterfall Atrous Spatial Pooling (WASP) module for single person, multi-person top-down and multi-person bottom-up pose estimation. The WASP module is the inspiration for the waterfall transformer module in WTPose, as it significantly increases the multi-scale representations and field-of-view (FOV) of the network and extracts features with a greater amount of contextual information, resulting in more precise pose estimates without the need for post-processing.

\subsection{Vision Transformers for Pose Estimation}

There is a recent surge of interest in models that employ transformer architectures for human pose estimation~\cite{TransPose_Yang_2021_ICCV},~\cite{TokenPose_Li_2021_ICCV},~\cite{9157545},~\cite{PoseFormer_zheng20213d},~\cite{HRFormer_yuan2021hrformer},~\cite{xu2022vitpose}. 
In earlier works, a CNN backbone was used as a feature extractor and the transformer was treated as a superior decoder~\cite{TransPose_Yang_2021_ICCV},~\cite{TokenPose_Li_2021_ICCV}. The TransPose~\cite{TransPose_Yang_2021_ICCV} architecture combines the initial parts of CNN-based backbones to extract features from images and the standard transformer architecture~\cite{vaswani2017attention} to utilize attention layers for learning dependencies and predicting keypoints for 2D human pose estimation. However, TransPose has a limitation in modeling direct relationships between keypoints. TokenPose~\cite{TokenPose_Li_2021_ICCV} explicitly embeds each keypoint as a token and simultaneously learns both visual clues and constraint relations through self-attention interactions. The HRFormer~\cite{HRFormer_yuan2021hrformer} is inspired by HRNet~\cite{wang2020deep} and utilizes a multi-resolution parallel design.  It adopts convolution in the stem and first stage, followed by transformer blocks. The transformer blocks perform self-attention on non-overlapping partitioned feature maps and use 3x3 depth-wise convolution for cross-attention among the partitioned maps. ViTPose~\cite{xu2022vitpose} adopts the plain and non-hierarchical vision transformer~\cite{dosovitskiy2020image} as a backbone to extract feature maps. The architecture then employs either deconvolutional layers or a bilinear upsampling-based decoder for 2D pose estimation. PoseFormer~\cite{PoseFormer_zheng20213d} proposed a pure transformer-based architecture for 3D pose estimation, based on 2D pose sequences in video frames.

\begin{figure*}[t]
\centering
  \includegraphics[width=\linewidth]{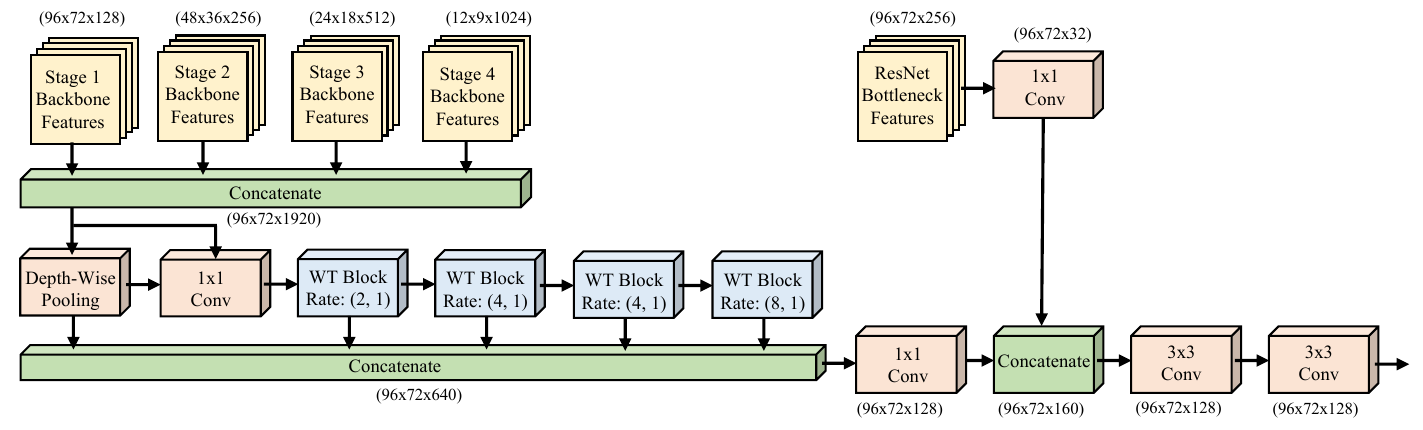}
  \caption{The proposed waterfall transformer module. The inputs are multi-scale feature maps from all four stages of the Swin backbone and low-level features from the ResNet bottleneck.
  The waterfall module creates a waterfall flow, initially processing the input and then creating a new branch. The  feature dimensions (spatial and channel dimensions) output by various blocks are shown in parentheses.}
    \label{fig:WTmodule}
\end{figure*}

\section{Waterfall Transformer}
\label{sec:waterfall_transformer}
The proposed waterfall transformer architecture, shown in Figure \ref{fig:WTarchitecture}, is a single-pass, end-to-end trainable network that incorporates a modified Swin transformer backbone and our transformer-based, multi-scale waterfall module for multi-person pose estimation. The patch partition layer in Swin ~\cite{liu2021swin} is replaced by two 3$\times$3 convolution (Stem) followed by the first residual block of ResNet-101~\cite{he2016deep_ResNet}, improving the feature representation of Swin. 

The processing pipeline of WTPose is illustrated in Figure \ref{fig:WTarchitecture}. The input image is fed to the transformer backbone which consists of our modified Swin transformer. The resulting multi-scale feature maps from multiple stages of Swin are processed by our waterfall transformer module (WTM) and are fed to the decoder to generate K heatmaps, one heatmap per joint. 
The multi-scale WTM maintains high resolution of feature maps and generates accurate predictions for both visible and occluded joints. 

The architecture of our waterfall transformer module is presented in Figure \ref{fig:WTmodule}. The WTM takes inspiration from the Disentangled Waterfall Atrous Spatial Pooling (D-WASP) module~\cite{Artacho_WACV2023},~\cite{BAPose_Sensor_s23073725}, which utilizes atrous blocks and the waterfall architecture to enhance the multi-scale representations. 
However, unlike D-WASP~\cite{Artacho_WACV2023},~\cite{BAPose_Sensor_s23073725}, which expands the FOV through atrous convolution, our proposed approach employs a dilated neighborhood attention transformer block to expand the FOV. The dilated transformer is built on the DiNAT~\cite{DiNAT_hassani2022dilated} architecture, featuring both dilated and non-dilated neighbouring attention. The dilated neighborhood attention expands the local receptive fields by increasing the dilation rates and performs sparse global attention. On the other hand, the non-dilated neighborhood attention confines self-attention of each pixel to its nearest neighbors.

To address contextual and spatial information loss resulting from the hierarchical backbone structure, the WTM processes multi-scale feature maps from all four stages of the Swin backbone through waterfall branches. 
The WTM module first performs upsampling operation using bilinear interpolation on the low-resolution feature maps from backbone stages 2, 3, and 4, to match them with the high-resolution feature maps from stage 1, and then combines all the feature maps to generate multi-scale feature representations for enhanced joint estimation. The multi-scale feature representation is then processed with 1$\times$1 convolutions to reduce the channel size to 128.
The concatenation (represented by the summation operator) of  the feature maps is $g_{0} = \sum_{i=1}^4(f_{i})$
where $f_{i}$ represents the feature maps from the Swin backbone and the index $i$ =1,2,3,4 indicates the Swin stages. The output after channel reduction is $ z_{0} = W_{1}\circledast g_{0}$
where $W_{1}$ denotes the 1$\times$1 convolution kernel and $\circledast$ represents convolution.

The output feature maps $z_{0}$ are then fed into the  waterfall transformer blocks (WTB) which expand the FOV by performing a filtering cascade at increasing rates. Each WTB contains two types of attention, dilated multi-head neighborhood self-attention (D-MHSA) followed by multi-layer perceptron (MLP) to capture global context, and non-dilated multi-head neighborhood self-attention (N-MHSA) followed by MLP to capture local context. 
\begin{align*}
\hat{z}_{l} &= \text{D-MHSA}(\text{LN}(z_{l-1})) + z_{l-1}, \notag \\
z_{l} &= \text{MLP}(\text{LN}(\hat{z}_{l})) + \hat{z}_{l}, \notag \\
\hat{z}_{l+1} &= \text{N-MHSA}(\text{LN}(z_{l})) + z_{l}, \notag \\
z_{l+1} &= \text{MLP}(\text{LN}(\hat{z}_{l+1})) + \hat{z}_{l+1}
\end{align*}
where $\hat{z}_{l}$ and $z_{l}$ denote the output features of the MHSA modules and MLP module for block $l$, respectively; D-MHSA and N-MSHA denote the multi-head self-attention based on dilated and non-dilated window, respectively.  

The waterfall module is designed to create a waterfall flow, initially processing the input and then creating a new branch. The WTM goes beyond the cascade approach by combining all streams from all its WTB branches and the depth-wise pooling (DWP) layer from the multi-scale representation. 
\vspace{-0.25in}

\begin{align}
    f_{Waterfall} = W_{1}\circledast(\sum_{i=1}^4(z_{i}) + \text{DWP}(g_{0}) \\
    f_{maps} = W_{3}\circledast(W_{3}\circledast(W_{1}\circledast f_{LLF} + f_{Waterfall})) 
\end{align}
\noindent 
where, summation denotes concatenation, 
$f_{LLF}$ are the low-level features from ResNet bottleneck, and $W_{1}$ denotes 1$\times$1 convolution, and $W_{3}$ denotes 3$\times$3 convolution with kernel size of 3 and strides of 1.

\begin{table*}
\small
\begin{center}
\begin{tabular}{|c|c|c|c|c|c|c|c|c|c|}
\hline
Method&Input Size&Params (M)&AP&$AP^{50}$&$AP^{75}$&$AP^{M}$&$AP^{L}$&AR\\
\hline\hline
HRNet-W32~\cite{sun2019deep}&384$\times$288&28.5&75.8&90.6&82.7&71.9&82.8&81.0\\
HRNet-W48~\cite{sun2019deep}&384$\times$288&63.6&76.3&90.8&82.9&72.3&83.4&81.2\\
Swin-B~\cite{mmpose2020}&384$\times$288&88.0&75.9&91.0&83.2&71.7&83.0&81.1\\
Swin-L~\cite{mmpose2020}&384$\times$288&197.0&76.3&\textbf{91.2}&83.0&72.1&83.5&81.4\\
ViTPose-B~\cite{xu2022vitpose}&384$\times$288&86.0&76.9&-&-&-&-&81.9\\

\textbf{WTPose}&384$\times$288&89.3&\textbf{77.1}&91.1&\textbf{84.1}&\textbf{73.4}&\textbf{83.9}&\textbf{82.0}\\
\hline
\end{tabular}
\end{center}
\caption{WTPose results and comparison on the COCO validation dataset.}
\label{tab:COCOval}
\end{table*}

\section{Experiments}
\label{sec:experiments}
We performed multi-person pose estimation experiments on Common Objects in Context (COCO) ~\cite{Lin2014MicrosoftCC}. The COCO dataset ~\cite{Lin2014MicrosoftCC} is composed of over 200K images in the wild and contains 250K instances of the person class. We train WTPose on COCO train 2017, with 57K images and 150K person instances, and validate on the val 2017 set containing 5K images. The labelled pose contains 17 keypoints. 

We adopt Object Keypoint Similarity (OKS)~\cite{Lin2014MicrosoftCC} to evaluate our model. Following the evaluation framework set by ~\cite{Lin2014MicrosoftCC}, we report OKS as the Average Precsion (AP) for the IOUs for all instances between 0.5 and 0.9 (AP), at 0.5 ($AP^{50}$) and 0.75 ($AP^{75}$), as well as instances of medium ($AP^M$) and large size ($AP^L$). We also report the Average Recall between 0.5 and 0.95 (AR).


We utilized the Swin Base (Swin-B) transformer as the backbone, initializing it with pre-trained weights from ~\cite{liu2021swin}. The default Swin-B architecture was implemented with the window size of 7. For the WTM module, we experimented with various rates of dilation and discovered that alternating between larger receptive field from dilated window and a small receptive field from non-dilated window results in improved prediction. We set the dilation rate for the WTB blocks  to (2,1), (4,1), (4,1), (8,1), with window size of 7.

Our models were trained on 4 A100 GPUs using the mmpose codebase~\cite{mmpose2020},  with a batch size of 32. We used the default training setting in mmpose to train WTPose, and employed the AdamW~\cite{AdamW_reddi2019convergence} optimizer with a learning rate of 5e-4. Our models were trained for 210 epochs, with a learning rate decay by 10 at the 170$^{th}$ and 200$^{th}$ epoch.

\begin{table}
\small
\begin{tabular}{|c|c|c|c|c|c|c|}
\hline
BN&Model&Waterfall&Dilation &AP&AR\\
\hline\hline
-&Swin-B&-&-&75.9&81.1\\
-&Swin-B&WTM&2,4,4,8&76.1&81.3\\
-&Swin-B&WTM&2,1,4,1,4,1,8,1&76.5&81.4\\
\checkmark&Swin-B&WTM&2,1,4,1,4,1,8,1&\textbf{77.1}&\textbf{82.0}\\
\hline
\end{tabular}
\caption{Results using different versions of WTPose on MS COCO validation dataset. All the models use Swin-B transformer as backbone with input image size of 384$\times$288. BN represents the Stem + ResNet bottleneck added at the initial layer of Swin and Waterfall indicates the use of the waterfall transformer module.
}
\label{tab:ablation}
\end{table}


\subsection{Experimental results on the COCO dataset}
We performed training and testing on the COCO dataset and compared WTPose with the Swin framework, as shown in Table 1. Our  WTPose model is 1.3M parameters larger than Swin-B, and it outperforms Swin-B in terms of average precision and average recall by 1.2\% and 0.9\%, respectively. Compared to Swin-L, our WTPose is about 54\% smaller and still outperforms Swin-L by 0.8\% and 0.6\% in terms of average precision and average recall. The waterfall transformer module improves the feature maps, increasing the accuracy of keypoint detection.

We performed ablation studies to investigate individual components of WTPose. Table 2 shows results with various configurations using the Swin-B backbone and an input image size of 384$\times$288. 
We set the window size to 7$\times$7, assign 8 heads for each attention layer, and select the dilation rates of 2, 4, 4, 8 to increase the size of the receptive fields at different WTB blocks.
The sizes of the receptive fields for dilation rates 1, 2, 4, and 8 are 7$\times$7, 13$\times$13, 25$\times$25, and 49$\times$49, respectively.  
First, we experimented with dilation rates of 2, 4, 4, 8 for each WTB, which involved performing a single dilated multi-head self-attention at each WTB with the assigned dilation rate. Next, we used both dilated and non-dilated multi-head self-attention mechanisms for each WTB and set the dilation rates as (2, 1), (4, 1), (4, 1), (8, 1) for each WT block.
Our main observations are (i) 
integrating the waterfall transformer module with the modified Swin backbone improves the feature representations, and
(ii) adding a Stem and ResNet Bottleneck at the start of the Swin-B further enhances the backbone's capability. 

\section{Conclusion}
We presented a Waterfall Transformer framework for multi-person pose estimation. WTPose incorporates our waterfall transformer module, which processes feature maps from various stages of the Swin backbone, followed by a cascade of dilated and non-dilated attention blocks to increase the receptive field and capture both local and global context. WTPose, with a modified Swin-B backbone and the waterfall transformer module, achieves improved performance over other Swin models.

\section{Acknowledgements}
This research was supported in part by the National Science Foundation grant $\sharp$1749376.


\end{document}